\documentclass[review,12pt]{elsarticle}

\usepackage{hyperref}
\usepackage{amssymb}
\usepackage{amsmath}
\usepackage{booktabs}
\usepackage{bm}
\newtheorem{theorem}{Theorem}
\newtheorem{lemma}{Lemma}

\newtheorem{Corollary}{Corollary}
\journal{ArXiv}
\begin{document}

\begin{frontmatter}



\title{Parallel ensemble methods for causal direction inference}


\author{Yulai Zhang$^{a}$, Jiachen Wang$^{a}$, Gang Cen$^{a}$, Guiming Luo$^b$ }
\address[a]{School of Information Technology and Electronics Engineering, \\ Zhejiang University of Science and Technology, Hangzhou, China, 310023, \\ zhangyulai@zust.edu.cn, jcwang.zust@foxmail.com, gcen@zust.edu.cn}
\address[b]{School of Software, Tsinghua University, Beijing, China, 100084, gluo@tsinghua.edu.cn}

\begin{abstract}
Inferring the causal direction between two variables from their observation data is one of the most fundamental and challenging topics in data science. A causal direction inference algorithm maps the observation data into a binary value which represents either $x$ causes $y$ or $y$ causes $x$. The nature of these algorithms makes the results unstable with the change of data points. Therefore the accuracy of the causal direction inference can be improved significantly by using parallel ensemble frameworks. In this paper, new causal direction inference algorithms based on several ways of parallel ensemble are proposed. Theoretical analyses on accuracy rates are given. Experiments are done on both of the artificial data sets and the real world data sets. The accuracy performances of the methods and their computational efficiencies in parallel computing environment are demonstrated.

\end{abstract}



\end{frontmatter}


\section{Introduction}

In many research fields, scientists do experiments to determine the causal relationships between variables. But in disciplines such as ecology and economics, the values of the variables can only be observed passively rather than be controlled by scientific experiments.
As a result, causal relationships have to be obtained by leveraging the data. In the early researches in \cite{pearl:2008}\cite{spirtes2000causation}, graphic models are constructed by calculating the conditional independence among at least three variables. 
However, inferring the causal direction between two variables is a tougher and more challenging task over years since the graphs represent $x$ causes $y$ and  $y$ causes $x$ determine the same Markov equivalence class \cite{gillispie2001enumerating}.

In the past decade, a family of methods based on the functional causal models (FCM) \cite{zhang2016estimation} are developed to deal with the problem of causal direction inference \cite{Mooij2016Distinguishing}.
%In the resent years,  to figure out the observed data $y = f(x)$ or $x = g(y)$. 
For the causal relationship represented by $y = f(x)$, the very basic idea of the FCM based methods lies in the fact that, the effect variable's probability density function $p_y$ should be more related with the transfer function $f$, than the cause variable's probability density function $p_x$. Many models and methods are proposed in the recent years by using this rule, either implicitly or explicitly, such as the ANM (Additive Noise Models) in \cite{peters2011causal}\cite{peters2014causal}, the PNL (Post Non-liear) in \cite{zhang2016estimation}\cite{Zhang2009On} and the IGCI (Information Geometric Causal Inference) in \cite{Mooij2016Distinguishing} and \cite{janzing2014justifying}. More recently, the RECI (Regression Error Causal Inference) method in \cite{bloebaum2018cause} uses the regression errors to determine the causal direction.  In \cite{goudet2018learning}, deep neural models are leveraged to fit the data. And in the Slope method in \cite{marx2019telling}, the minimum description lengths of the regression models are taken into consideration. To this day, causal direction inference remains a research hot spot in the data science community \cite{janzing2019cause}. %%%%%%%%%%%%%%%%%%%%

The causal direction inference algorithms can be taken as a map from the observation data to a binary value. This is quite different from the model based classification and regression algorithms commonly appeared in the machine learning and data mining literature. These methods usually map the training data into a set of model parameters $\theta$ in the real number domain $\mathcal{R}^{|\theta|}$. The result of $\theta$ will not be changed dramatically by adding or deleting one or a few data points in the training data set. This is not the case in causal direction inference. The result of the causal direction can be changed simply by adding or deleting one sample point, just as depicted in the Figure 1. This is an unacceptable feature for the task of causal inference. The causal relationships are supposed to be stable and rarely changed over time. In addition, %this is one of the major reasons for the loss of accuracy, which is more concerned with. An 
an opposite causal direction will lead to totally different judgments and decisions when applied in large and complex systems in many application fields.   
	
	%Unlike the other learning algorithms. The result of causal direction inference is 1 bit. The causal direction algorithm function as a map from data space $R^2-\>bool$ to a boolean variable.   This type of methods are called unstable learners in the machine learning literature. One of the most effective way to improve the performance of these methods is the parallel ensemble.
In this work, parallel ensemble \cite{Zhou2012Ensemble} based methods for causal direction inference are proposed to improve the accuracy and stability of the causal direction decisions. Ensemble method is a research hot spot in the recent years to deal with the growing data volumes\cite{ALJARRAH201587}. Many conventional methods can be greatly improved by using an ensemble framework. For instance, gradient boost decision tree model \cite{ke2017lightgbm}, which is the sequential ensemble version of the conventional decision tree \cite{Quinlan1986Induction} is very popular nowadays.   
In a parallel ensemble method, the final result is obtained by aggregating the results of every base learners. If the base learners are capable of performing better than random guesses, the final result of their ensemble  are guaranteed to be better than the results of individual base learners. The only problem is that in parallel ensemble, the base learners use subsets of the original data set, that will reduce the accuracy rates of the base learners. Consequently, the trade off between the number of base learners and the re-sampling rates should be carefully handled. In addition, the parallel ensemble mechanism may make the time complexity increase by almost $T$ times, where $T$ is the number of the base learners. Fortunately, the algorithm is very easy to be deployed in parallel computing environments. 

There are two contributions in this paper. First, a parallel ensemble framework for causal direction inference is proposed. Second, the accuracy of the framework is analyzed when IGCI is used as the base method. So the optimal choices of the sub set size and the number of base learners are investigated based on this accuracy analysis. The idea of parallel ensemble has not been used in the field of causal direction inference. We find two empirical research papers in social science\cite{sliva2017modeling} and economics \cite{athey2019ensemble}, where they use the idea of simple ensemble to determine the causal relationships by collecting the results from several different causal inference methods. In this paper, our base learners use the same algorithms, so the accuracy can be analyzed theoretically. 

In the rest of this paper, the preliminaries of the causal criterion of the base learners are given in the Section 2. The novel algorithms and the corresponding accuracy analysis are proposed in the Section 3. Experiments are demonstrated in the Section 4. And the conclusions are given at the end of the article. 

\section{Preliminaries}

When two variables $x$ and $y$ are statistically related, there are five possible scenarios \cite{Mooij2016Distinguishing}.
i) $x$ causes $y$; ii) $y$ causes $x$; iii) dependence and feedback;  iv) hidden confounder (which means $x$ and $y$ have a common cause); v) selection bias (which means $x$ and $y$ have a common effect, and they are observed conditionally on this variable). 
Note that the relationship in the real world may be more sophisticated, for instances, the feedback and hidden confounder may exist at the same time (a combination of scenarios iii and iv). We would like to simplified the discussion by considering only the scenarios i and ii. That is to say, we assumed that there is no feedback, no hidden confounders and no selection bias in the observation data.

\subsection{Causal criterion in the base method}\label{secigci}
In causal direction inference problems, the values of the cause and the effect are observed and stored in the data set $D =\{\mathbf{x},\mathbf{y}\}$, where $\mathbf{x} = \{x_1,x_2,...,x_m\}$ and  $\mathbf{y} = \{y_1,y_2,...,y_m\}$, $m$ is the number of sample pairs. 
In the context of functional casual models, if $x$ is the cause variable and $y$ is the effect variable, their relationship can be expressed as $y = f(x)$. The observations may be contaminated by noise, so the data should be expressed as $x = \hat{x}+\epsilon_x$, $y = \hat{y}+\epsilon_y$ and $\hat{y} = f(\hat{x})$. Note that many causal inference methods use the additive noise assumption to tell cause from effect such as ANM, PNL and LinGAM. In some other methods such as IGCI, observation noise can also be part of the reason for their unstable performances.

Similarly, if the truth is $y$ causes $x$, the relationship can be described as $x = g(y)$. The task of the causal direction inference is to find out the correct model that produces the data set $D$.  From the perspective of regression method, $f$ and $g$ can fit the data points equally well. So it is an uneasy task to tell the correct direction. By the way, $f$ or $g$ can have explicit formulas, or they can also be described by non-parametric models such as Gaussian process.

The IGCI method mentioned in the introduction will be chosen as the base learners in this work. Not only because it achieves one of the best results on the benchmark data sets, but also because that no regression models are used to fit the data in this method. The results of the regression based methods such as ANM may be influenced by the problems such as under fitting or over fitting. However, the discussion of this paper will be focus on the effect of the ensemble. 
In addition, the framework proposed in this work can also use other causal direction inference methods as the base learners.

%Note that the values of $x$ and $y$ can be disturbed by noise, which can also be explained as the minor factors of the effect variable. In addition, $f$ or $g$ may have explicit formulas, or they can be non-parametric models such as Gaussian process. 

For any nonlinear function $f$, let $y$ be produced by $y = f(x)$ and $g$ is the inverse function of $f$. The probability density of the effect variable $p_y$ can be expressed as $p_y(y) = f(g(y))\cdot g'(y)$. So the method IGCI assumes that $p_y$ and $g'(y)$ should be more related with each other than $p_x$ and $f'(x)$. This asymmetric property can be used to infer the causal direction between $x$ and $y$. 
%In fact the proposed ensemble framework can be effective to any base learners whose accuracy is significantly better than a random guess ($>50\%$).

Suppose that $x$ causes $y$, the causal criterion of IGCI can be formally expressed as
\begin{equation}\label{igci0}
\int_{0}^{1}log|f'(x)|p(x)dx < \int_{0}^{1}log|g'(y)|p(y)dy
\end{equation}
%$$\int_{0}^{1}log|f'(x)|p(x)dx < \int_{0}^{1}log|f'(y)|p(y)dy$$
Thus the causal direction of $x$ and $y$ can be obtained by comparing $e_x$ and $e_y$, who are the mathematical expectation of $log|f'(x)|$ and $log|f'(y)|$ respectively. 
\begin{equation}\label{igci1}
e_x = \frac{1}{m-1}\sum_{i=1}^{m-1}log|\frac{y_{i+1}-y_i}{x_{i+1}-x_i}|
\end{equation}
\begin{equation}\label{igci2}
e_y = \frac{1}{m-1}\sum_{j=1}^{m-1}log|\frac{x_{j+1}-x_j}{y_{j+1}-y_j}|
\end{equation}
%$$e_x = \frac{1}{m-1}\sum_{i=1}^{m-1}log|\frac{y_{i+1}-y_i}{x_{i+1}-x_i}| $$
%$$e_y = \frac{1}{m-1}\sum_{i=1}^{m-1}log|\frac{x_{i+1}-x_i}{y_{i+1}-y_i}| $$
The data pairs in \eqref{igci1} are sorted by $x$, and the data pairs in \eqref{igci2} are sorted by $y$, so the data pairs are in different orders when calculating $e_x$ and $e_y$.
$e_x < e_y$ indicates that $x$ causes $y$, and $e_y < e_x$ indicates that $y$ causes $x$.

%The IGCI method is widely used and makes the best result in the benchmark data set in \cite{Mooij2016Distinguishing}.
%
%The criterion in IGCI can be fomulated as 
%
%$$S(P_X) = \psi(m) - \psi(1) = \frac{1}{m-1}\sum_{i=1}^{m-1}log\|x_{i+1}-x_i\|$$    
%
%The 
%$$C_{x->y} = \frac{1}{m-1}\sum_{i,j=1}^{m-1}log\|\frac{x_{i+1}-x_i}{y_{j+1}-y_j}\|  $$

Note that if there are repeated values in the data set, some of the denominators in \eqref{igci1} and \eqref{igci2} will be zero. So the following procedures are used instead. First record the number of elements that have the same value with $x_i$ as $n_i$, and then remove the repeated elements. Let the new data sequence be $\hat{\mathbf{x}} = \{\hat{x}_1,\hat{x}_2,...,\hat{x}_{\hat{m}}\}$. The $e_x$ in \eqref{igci1} can be calculated as
\begin{equation}\label{igci_rep1}
    \hat{e}_x = \frac{1}{\sum_{i=1}^{\hat{m}_x-1}n_i}\sum_{i = 1}^{\hat{m}_x-1} n_ilog\frac{|\hat{y}_{i+1} - \hat{y}_i|}{|\hat{x}_{i+1} - \hat{x}_i|}.
\end{equation}
Similarly, record the number of elements that have the same value with $y_j$ as $n_j$ and remove the repeated elements, the $e_y$ in \eqref{igci2} can be calculated as
\begin{equation}\label{igci_rep2}
\hat{e}_y = \frac{1}{\sum_{j=1}^{\hat{m}_y-1}n_j}\sum_{j = 1}^{\hat{m}_y-1} n_jlog\frac{|\hat{x}_{j+1} - \hat{x}_j|}{|\hat{y}_{j+1} - \hat{y}_j|}.
\end{equation}
\subsection{The unstable performance of the base method}
\begin{figure}[!htb]
	\centering
	\includegraphics[width=285pt, bb=0 0 475 321]{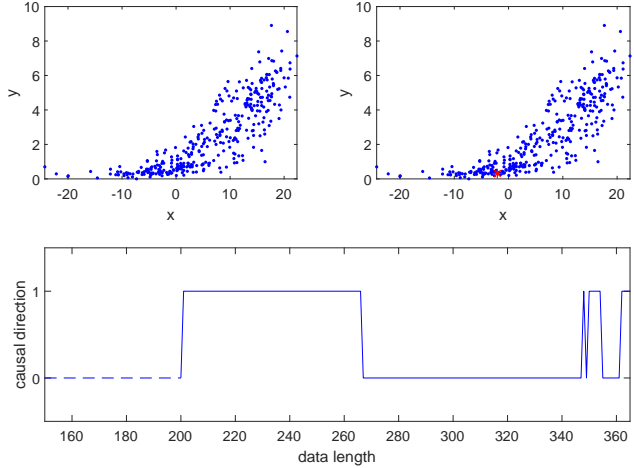}\\
		\caption{Unstable performance of the base method. The initial data set size is $200$ and a pair of $x$ and $y$ is added at a time to expand the data set. Top left figure: scatter figure of $x$ and $y$ at step $348$; Top right figure: scatter figure of $x$ and $y$ at step $349$, where the new data point is marked in red. Bottom figure: The horizontal ordinate represents the length of the data sets. The vertical ordinate represents the causal direction determined by the base method. 1 denotes $x$ causes $y$, which is correct, and 0 means $y$ causes $x$.}\label{unstable}
\end{figure}

Just as mentioned in the induction part, the results of the causal direction inference are rather unstable. An example is given here to raise the issue. %The results of the IGCI algorithm on a data set from the fluxnet \cite{yu2019anticipating} are depicted in the Figure \ref{unstable}. 
The data pairs are collected by Fluxnet\cite{yu2019anticipating}. $x$ denotes the night temperature and $y$ is the $CO_2$ flux at night at the same place. The respiration of plants will be stronger at higher temperature. So here $x$ causes $y$. One pair of data is recorded each day. So there are 365 data pairs in this data set. The results of the IGCI method are depicted in the Figure \ref{unstable}. 
The experiments are done simultaneously with the data collecting process. The first experiment is done with the data pairs collected from the very beginning to the 200th day and the second experiment is performed with the data pairs collected up to the 201th day, etc. The causal decisions of the base method are depicted in the Figure \ref{unstable}. It shows that the results of the base method are rather unstable and almost half of the results are incorrect. The algorithm's output can be changed simply by adding one data pair. Specifically speaking , a new data point breaks the sorted sequences of $x$ and $y$ in \eqref{igci1} and \eqref{igci2}. If the new slopes in one direction are sufficiently larger than the other direction, then the sign of the result $e_x-e_y$ may be changed. This is the source of instability in the IGCI method. It is partly due to the short of data samples but we believe that the stochastic nature of the specific data set plays a more important role. This is an undesirable feature for the task of causal direction inference. Performances like this are also reported in other applications\cite{zhang2018causal}.

\section{Parallel ensemble methods for causal direction inference}\label{sec3}

\subsection{Parallel Ensemble Causal Direction Inference}
If the correct rate of a causal direction inference method is higher than $50\%$, a more stable and probably correct result can be anticipated by combining multiple answers of the model using the majority vote strategy. A parallel ensemble framework is proposed in this section. The original data set is re-sampled to generate a number of sub data sets for base learners. The final decision is the majority vote of the base learners. The algorithm is described in the Table \ref{peci}. $\mathbf{x}$ and $\mathbf{y}$ are the original data with length $m$. $T$ is the number of the base learners. $k$ is the length of the re-sampled data sets, and $k<m$. The votes are recorded in the vector $\mathbf{r}$, $1$ for one causal direction and $-1$ for the other. The finally decision can be obtained by $\sum r_t$. 

Note that if $k$ is too close to $m$, the diversity of the base learners may be insufficient. They tends to give the same answer. So the performance of the new algorithm will be very close to that of the base learner on the original data set. Otherwise if $k$ is much smaller than $m$, the accuracy of the base learners will drop with $k$. 
An optimal value of k should be determined by taking the above two issues into consideration. On the other hand, it is easy to see that the accuracy will increase with $T$. We would like to investigate the influences of the parameter settings in the next subsection. 
%\begin{center}
%\renewcommand\tablename{算法}
\begin{table}[ht]%\large
	\caption{Parallel Ensemble Causal Direction Inference} \centering
	\begin{tabular}{p{0.8\textwidth}c}
		\toprule
		\textbf{PECI(Parallel Ensemble Causal direction Inference)}&  \\
		\midrule
		$\mathbf{Input}$:  $\mathbf{x}$, $\mathbf{y}$, $k$, $T$\\%，$x_2$ \\
		$\mathbf{Output}$: causal direction \\
		
		1 \ \ \ \ Let $\mathbf{r}$ be an $T\times1$ zero vector; \\
		2 \ \ \ \ for t from 1 to T: \\
		3 \ \ \ \ \ \ \ \ \ \ Sample $k$ data pairs randomly from $\mathbf{x}$, $\mathbf{y}$ as  $\mathbf{x^{(t)}},\mathbf{y^{(t)}}$;\\
		4 \ \ \ \ \ \ \ \ \ Sort data by $\mathbf{x^{(t)}}$ and calculate $e_x$ by Eq.\eqref{igci_rep1}\\
		5 \ \ \ \ \ \ \ \ \ Sort data by $\mathbf{y^{(t)}}$ and calculate $e_y$ by Eq.\eqref{igci_rep2}\\
		6 \ \ \ \ \ \ \ \ \ if $e_x<e_y$ let  $r_t = 1$\\		
		7 \ \ \ \ \ \ \ \ \ else if $e_x>e_y$ let  $r_t = -1$\\	
		8 \ \ \ \ end for\\
		9 \ \ \ \ if $\sum_t r_t>0$ output causal direction $x \rightarrow y$\\
	    10 \ \ \ else if $\sum_t r_t<0$ output causal direction $y \rightarrow x$\\
	    11 \ \ \ else output unknown\\
		\bottomrule
	\end{tabular}\label{peci}
\end{table}

\subsection{Accuracy Analysis of PECI}
In the base method, the logarithmic slope $e_i = log|\frac{y_{i+1}-y_i}{x_{i+1}-x_i}|$ can be taken as a random variable with arbitrary distribution. Then according to the central limitation theorem, $e = \sum e_i$ approximates to the Gaussian distribution. This is written as Lemma 1 and the error rate of the base method can be consequently obtained in the Lemma 2. The upper bound of the error rate of the proposed method can be derived in Theorem 1. 

\begin{lemma}\label{lem1}
	Let the logarithmic slopes $e_i = log|\frac{y_{i+1}-y_{i}}{x_{i+1}-x_{i}}|$ in the equation \eqref{igci1} be a set of random variables with independent identical distribution $p_e$ whose mathematical expectation and variance are $\mu_x$ and $\sigma_x$ respectively. If the number of data points is sufficiently large, the variable $e_x = \frac{1}{m-1}\sum_i e_i$ in the equation \eqref{igci1} can be considered as a random variable with normal distribution $\mathcal{N}(\mu_x, \frac{\sigma_x}{\sqrt{m-1}})$.%, where $\mu_x = \frac{1}{m-1}\sum_{k}\mu_k$, $\sigma_x = \frac{1}{m-1}\sum_{k} \sigma_k$.   
		
\end{lemma}
Lemma \ref{lem1} can be derived from the central limitation theorem directly.
Identically, in the other direction, the variable $e_y$ in \eqref{igci2} can also be regarded as a random variable with normal distribution $\mathcal{N}(\mu_y, \frac{\sigma_y}{\sqrt{m-1}})$.% $m$ is the number of data pairs.

\begin{lemma}\label{lem2}
	Under the assumptions of Lemma \ref{lem1}, the error rate of the base method IGCI described the section \ref{secigci} can be expressed as
	\begin{equation}\label{er1}
	\varepsilon_{m} = \frac{1}{2}(1-\mathrm{erf}(\sqrt{m-1}\cdot\frac{\mu}{\sqrt{2\sigma^2}})
	\end{equation}
		where $\sigma^2 = \sigma_x^2+\sigma_y^2$ and $\mathrm{erf}(\cdot)$ is the Gaussian error function which is defined as $\mathrm{erf}(x) = \frac{2}{\sqrt{\pi}}\int_0^{x}e^{-t^2}dt$. If the truth is $x$ causes $y$, $\mu = \mu_y - \mu_x$, otherwise  $\mu = \mu_x - \mu_y$.
\end{lemma}
Lemma \ref{lem2} can be simply derived by the fact that the based method will give correct results when $e_y - e_x > 0$, so the error rate $\varepsilon_{m}$ can be obtained by calculating $p(e_y - e_x \le 0)$ (for the case $x$ causes $y$). The subscript $m$ denotes the length of the data set. The error rate will be decreased by collecting more data pairs when $\mu>0$. Note that if $\mu < 0$, the output of the $\mathrm{erf}(\cdot)$ function will be negative, and then the error rate of the base method will be greater than $1/2$, which means the correct rate can be worse than random guesses and collecting more data pairs does not help improving the performances.

%%%%%%%%%%%%%%%%%%%%%%%%%%%The following text is from wiki
%The inequality states that the probability that the estimated and true values differ by more than t is bounded by e−2nt2. Symmetrically, the inequality is also valid for another side of the difference:
%%%%%%%%%%%%%%%%%%%%%%%%%%%5

From Lemma 2 and The Hoeffding's inequality, the upper bound of the error rate of PECI can be derived. 

\begin{theorem}(The upper bound of the error rate of PECI.) Let $T$ be the number of ensemble tasks and $k$ be the re-sampling size of each ensemble task in the algorithm PECI. The upper bound of the error rate $\varepsilon_k^{(T)}$ can be obtained by %The upper bound of a simple vote parallel ensemble.
%Error rate of $m/N$ sample is $\varepsilon$, the task number is $N$, the final error rate with a simple voting straitigy is 
%\begin{equation}
%\varepsilon_k^{(T)} \le \mathrm{exp}(-\frac{2}{T}\mathrm{erf}^2(\frac{\sqrt{k-1}\mu}{\sqrt{2\sigma^2}}))
%\end{equation}
\begin{equation}\label{the}
\varepsilon_k^{(T)} \le \mathrm{exp}(-\frac{T}{2}\mathrm{erf}^2(c\sqrt{k-1}))
\end{equation}
where $c = \frac{\mu}{\sqrt{2\sigma^2}}$ in the Lemma 2.

\end{theorem}
The proof of theorem 1 is provided in the appendix. 
%\begin{Corollary}
%	content...
%\end{Corollary}
From the above theorem we can see that the upper bound decreases with $k$ and $T$. However, the choice is not that simple. The size of the sub data set should not be too small, sufficient number of data points should be sampled to keep the accuaracy of the base method. On the other hand, it should also not be too large, otherwise the sub data sets may be lack of diversity, which causes the results of the base method being the same. So we should find a appropriate $k$ between $0$ and $m$. It seems that we should let the value of $T$ to be as large as possible. However, the number of $T$ is limited by $T<C_m^k$, since ensemble with identical tasks is meaningless. From the above discussion we can see that the ensemble framework does not guarantee better results, unless the parameters are carefully selected. So the next question is to give the condition under which a better result can be guaranteed.

\begin{Corollary}
The ensemble algorithm PECI will have better results than its base method IGCI if the value of the parameter $k$ satisfies \eqref{coro1} and the value of the parameter $T$ satisfies \eqref{coro2},
\begin{equation}\label{coro1}
C_m^k \mathrm{erf}^2(c\sqrt{k-1}) > E_c(m)
\end{equation}
\begin{equation}\label{coro2}
T> \frac{E_c(m)}{\mathrm{erf}^2(c\sqrt{k-1})}
\end{equation}
where $E_c(m) = 2ln\frac{2}{1-erf(c\sqrt{m-1})}$, and it is determined by the nature of the data $c = \frac{\mu}{\sqrt{2\sigma^2}}$ and the number of data points $m$.	
\end{Corollary} 
The inequalities \eqref{coro1} and \eqref{coro2} can be easily obtained from Lemma 2 and Theorem 1. Note that the left hand side of the inequality \eqref{coro1} is nonmonotonic which means there is an optimal value for $k$. However, it is not a solid conclusion that the optimal value of $k$ is $k^* = \arg\max_k (C_m^k \mathrm{erf}^2(c\sqrt{k-1}))$. We would like to leave that to the further researches.

\subsection{Weighted PECI}

From the Hoeffding's inequality in the Lemma 3 in the appendix section we can see that the upper bound also holds when the results of the base estimators are weighted. 
The values of $|e_x - e_y|$ can be taken as the confidence of the causal direction decisions. So we can use them to formulate the weight of the votes to improve the PECI method. However, the absolute value of $e_x - e_y$ should be normalized to $[-1,1]$ to avoid the extreme outliers. The $tanh$ function in \eqref{tanh} and the transformation of the sigmoid function in \eqref{sig_w} are most frequently used in the machine learning literature. They can be represented as
\begin{equation}\label{sig_w}
    r  = \frac{1+e^{-(e_x-e_y)}}{1-e^{-(e_x-e_y)}}
\end{equation}
and
\begin{equation}\label{tanh}
   r = tanh(e_x-e_y).
\end{equation}
Note that $r$ can be either positive or negative. The final decisions can still be obtained by $\sum r_t$. The weighted algorithm is summarized in the Table \ref{peci_w}. The major difference with PECI lies in the Line 6.

\begin{table}[!ht]%\large
	\caption{Weighted Parallel Ensemble Causal Direction Inference} \centering
	\begin{tabular}{p{0.9\textwidth}c}
		\toprule
		\textbf{WPECI(Weight Parallel Ensemble Causal direction Inference)}&  \\
		\midrule
		$\mathbf{Input}$:  data pair $\mathbf{x}$, $\mathbf{y}$, $k$, $T$\\%，$x_2$ \\
		$\mathbf{Output}$: causal direction \\
		
		1 \ \ \ \ Let $\mathbf{r}$ be an $T\times1$ zero vector; \\
		2 \ \ \ \ for t from 1 to T: \\
		3 \ \ \ \ \ \ \ \ \ \ Sample $k$ data pairs randomly from $\mathbf{x}$, $\mathbf{y}$ as  $\mathbf{x^{(t)}},\mathbf{y^{(t)}}$;\\
		4 \ \ \ \ \ \ \ \ \ Sort $\mathbf{x^{(t)}}$ and calculate $e_x$ by Eq.\eqref{igci_rep1}\\
		5 \ \ \ \ \ \ \ \ \ Sort $\mathbf{y^{(t)}}$ and calculate $e_y$ by Eq.\eqref{igci_rep2}\\
		6 \ \ \ \ \ \ \ \ \ Calculate $r_t$ by Eq.\eqref{sig_w} or Eq. \eqref{tanh}\\		
		7 \ \ \ \ end for\\
		8 \ \ \ \ if $\sum_t r_t>0$ output causal direction $x \rightarrow y$\\
		9 \ \ \ \ else if $\sum_t r_t<0$ output causal direction $y \rightarrow x$\\
		10 \ \ \ else output unknown\\
		\bottomrule
	\end{tabular}\label{peci_w}
\end{table}

%The novelty of the proposed method lies in the re-sampling procedure and the way to   

\section{Experiments}

\subsection{Experiments on artificial data}
There are two groups of experiments in this subsection. In the first group, the data sets are produced by functions with explicit analytical equations. And in the second group, the data sets are produced by Gaussian process model, where the input and output of the model cannot be expressed by any explicit analytical equations. The second group of data is actually proposed by \cite{Mooij2016Distinguishing} as a benchmark for causal direction inference.

\subsubsection{Data from explicit analytical equation}
In this part, the data pairs are produced by an explicit formula $\hat{y} = exp(\hat{x})$ with additive noise $y = \hat{y}+\epsilon_y$ and $x = \hat{x}+\epsilon_x$. The values of $\hat{x}$ are generated by Gaussian distribution with zero mean and unit variance. $\epsilon_x$ and $\epsilon_y$ are zero mean Gaussian white noise whose variances are set to be $40$. The data points are normalized by $\frac{x-min(x)}{x-max(x)}$ to be within $[0,1]$.
There are 2000 data pairs in each simulation. And the simulations are repeated 10000 times. The values of accuracy are calculated based on the numbers of correct results. Note that the additive noise can also be taken as the unobserved latent variables. And the same 10000 data sets are used for different k and T, so the result of the based method is a horizontal line.
\begin{figure}[!htb]
	\centering
	\includegraphics[width=285pt, bb=0 0 420 315]{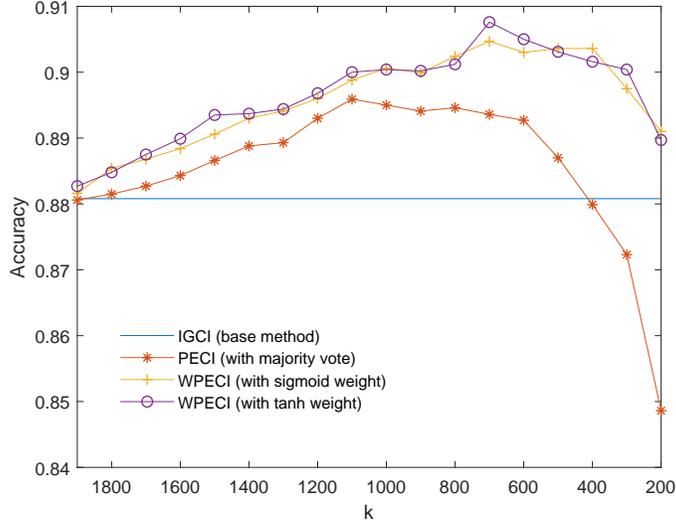}\\
	\caption{Performance of the methods under different $k$}\label{figure_k}
\end{figure}

In the experiments depicted in the Figure \ref{figure_k}, the number of tasks $T$ is set to be $100$, and the length of re-sampling $k$ ranges from $1900$ to $200$. From the result we can see that the when $k$ is close to the original data length $m=2000$, the accuracy rates of the based method and that of the ensemble method are close. This is because the data sets obtained by the re-sampling process are very close to the original data set when $k$ is very close to $m$. The accuracy rates of the ensemble methods increase while $k$ decreases from $1900$ to about $800$. After that the accuracy rates decrease with $k$. The performance of the PECI with majority vote is even worse than the base method for $k<400$. So this is actually caused by the decrease of the base learners' accuracy rates. The trade off between the accuracy rates of every single base learners and the diversity of them as a whole is clearly revealed by this picture. 

\begin{figure}[!htb]
	\centering
	\includegraphics[width=285pt, bb=0 0 420 315]{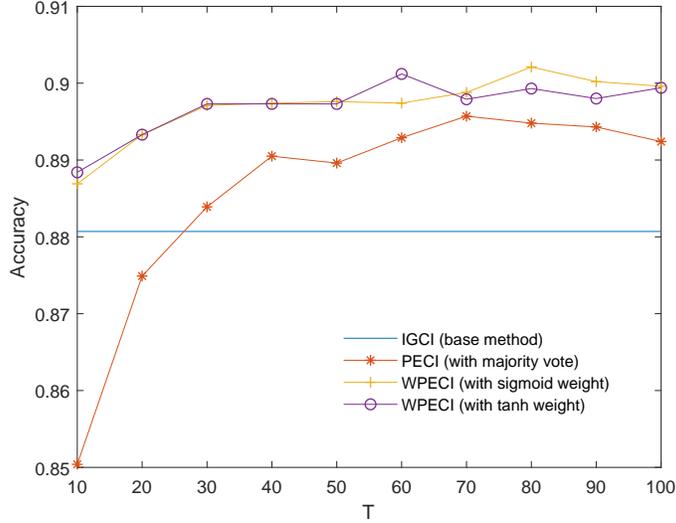}\\
	\caption{Performance of the methods under different $T$}\label{figure_T}
\end{figure}

In the experiments depicted in Figure \ref{figure_T}, $k$ is set to be $1000$, which equals $m/2$. The number of base learners $T$ ranges from $10$ to $100$. From the results we can see that the accuracy rates of the ensemble methods increase with the number of base learners. And the curve of PECI fits well with \eqref{the} in the Theorem 1. It is worth mentioning that, the value of $T$ cannot be arbitrarily large. It should be bounded by $C_m^k$, and if it is too close to this upper bound, there will be lack of diversity for the based learners, so computing resources will be wasted.

\begin{figure}[!htb]
	\centering
	\includegraphics[width=285pt, bb=0 0 420 315]{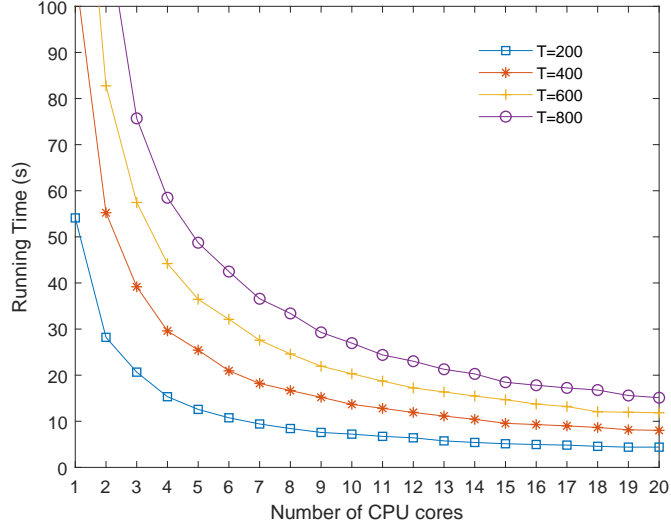}\\
	\caption{The time efficiency of PECI on parallel computing environment with different number of concurrences. }\label{figure_time_c}
\end{figure}

In Figure \ref{figure_time_c}, the running times of the PECI method on the parallel computing environment with different numbers of threads are depicted. The simulation is done on a workstation with Intel Xeon CPU E5-2650v3 2.3GHz. The parallel computing toolbox of Matlab\cite{sharma2009matlab} is used. The loop between Line 2 and Line 8 in the algorithm PECI in the Table \ref{peci} are set to be run in parallel. The running time decreases significantly with the increase of the CPU cores being used. The parallel ensemble method naturally fits the parallel computing environments.

\subsubsection{Data from Gaussian processes}
The data sets in this part are the artificial benchmark data sets in \cite{Mooij2016Distinguishing}. There are four groups of data sets who are produced by Gaussian process model with the additive noise. A brief description of the data generation is given here for the integrity of the article. This data set can be downloaded from the link provided in \cite{Mooij2016Distinguishing} and further details can also be found in its appendix. For SIM, SIM-ln and SIM-G, the values of $\mathbf{x}$ are sampled from the Gaussian process $\mathbf{x} \sim K_{\theta_1}(e_1)+\tau^2I$ and the values of $\mathbf{y}$ are sampled from the Gaussian process $\mathbf{y} \sim K_{\theta_2}((\mathbf{x},e_2))+\tau^2I$. $K_\theta$ is the Gram matrix of an RBF kernel function $k(\mathbf{x},\mathbf{x'})=\sum_i exp(-(x^{(i)}-x'^{(i)})^2/(\theta^{(i)})^2)$ with parameter $\theta$, the superscript $(i)$ represents the i th element of the vector. $\tau = 10e-4$ and $I$ is the Identity matrix. $e_1$, $e_2$ and $\theta_1$ and $\theta_2$ are also randomly sampled from Gaussian processes independently. The data sets in SIM-ln has relatively low noise. The distributions of the data sets in SIM-G are most similar to Gaussian distributions. The data sets in SIM-c are generated with comfounders. For SIM-c the values of $x$ and $y$ are sampled from  $\mathbf{x} \sim K_{\theta_3}((e_1,e_3))+\tau^2I$ and $\mathbf{y} \sim K_{\theta_4}((\mathbf{x},e_2,e_3))+\tau^2I$.

There are 100 identically distributed data sets in each group. And the data length of each single data set is $m = 1000$. $T$ is set to be $200$ and $k$ is $250$. 
The simulation results are recorded in the Table \ref{dgp}. 
\begin{center}
	\begin{table}[ht]\normalsize
		\caption{Comparison for the methods on the artificial benchmark data.} \centering
		\begin{tabular}{cccccc}  \hline
			%\toprule
			Data set & IGCI  & PECI with & WPECI with  & WPECI with\\
			Accuracy&base method&  majority vote& sigmoid weight&  tanh weight\\\hline
			%\midrule
			Sim-ln&62\%&63\%&65\% &62\%  \\
			Sim-G& 85\%& 91\% &90\%&89\%\\
			Sim-c &46\% &46\% &45\%&45\% \\
			Sim &42\% &40\% &39\%&39\%\\
			\hline
			
			%\bottomrule
		\end{tabular}\label{dgp}
	\end{table}
\end{center}
From the results in the Table \ref{dgp}, we can see that if the base method performs better than random guess, the accuracy of the PECI will be better than the base method. Otherwise if the base method performs worse than random guess, the result of PECI will be even worse. This is accord with the common sense of the ensemble methods.

\subsection{Simulations on the real world data}

The real world benchmark data repository proposed in \cite{Mooij2016Distinguishing} currently has 108 data sets from different disciplines and the number is ever increasing. 
In the simulations, we let the parameter $T=10000$ and let $k$ be changed with $m$ as $k= \lfloor 3m/4 \rfloor$ if $m \le 500$, $k = \lfloor m/2 \rfloor$ if $500<m\le 1000$, $k = \lfloor m/4 \rfloor$ if $1000<m\le2000$, $k = \lfloor m/8 \rfloor$ if $2000<m\le 10000$, and $k = \lfloor m/10 \rfloor$ if $m>10000$.
The above choice is based on two considerations. First, the size of the sub data sets should have enough data points in order to keep the accuracy of the base method. Second, more diversity can be obtained by smaller $k$ compared with $m$.
Beside IGCI, the other two recently proposed methods are also joined as the base method here to enrich the results. These two methods have to use a regression model to fit the data. For the method Slope\cite{marx2019telling}, the algorithm chooses the model with minimum description length among nine choices automatically as the regression model.   
And for the method RECI \cite{bloebaum2018cause}, nine-th order polynomial model is used as the regression model. 

\begin{center}
	\begin{table}[ht]\normalsize
		\caption{Comparison of different base method and their ensemble on real world data} \centering
		\begin{tabular}{ccccc}  \hline
			%\toprule
		    Name of the& Accuracy of & \multicolumn{3}{c}{Accuracy of ensemble method with}\\%Accuracy with & Accuracy with  & Accuracy with\\
			base method&base method&majority vote&sigmoid weight&tanh weight\\\hline
			%\midrule
			Slope \cite{marx2019telling}&59.2\%&60.1\%&61.1\% &61.1\%  \\
			IGCI& 65.7\%& 68.5\% &67.6\%&66.7\%\\
			RECI \cite{bloebaum2018cause}&75.0\% &80.6\% &80.6\%&80.6\% \\
			%Sim &42\% &40\% &39\%&39\%\\
			\hline			
			%\bottomrule
		\end{tabular}\label{dgp}
	\end{table}
\end{center}

The base method IGCI gives 71 correct causal directions in all the 108 data sets, and PECI with majority vote gives 74 correct answers, while WPECI with sigmoid weights and tanh weights give 73 and 72 correct answers respectively.
The base method Slope gives 64 correct causal directions, and the corresponding ensemble method with majority vote gives 65 correct answers. The correct numbers with sigmoid weights and tanh weights are both 66 respectively.
The base method RECI gives 81 correct results and its three ensemble counterparts all give 87 correct results. 

The results show again that if the base method has a higher accuracy, the performance of its ensemble version will give better results. For the ensemble methods with Slope and RECI, weighted algorithms outperform the algorithm with majority vote. But for the base method IGCI, it is not the case. We carefully examined the data set Number 20, on which the method with majority vote succeeded but the weighted algorithm failed. In the data set $x$ is the latitude and $y$ is the average temperature. The number of correct decisions are larger than the incorrect ones, but the absolute values of $e_x-e_y$ in the incorrect decisions are much larger than the correct ones. It dues to the fact that the stations with very close latitude can have very different temperature records. And the results are always very close to the edge $0$. Another reason we believe here is that the data set is relatively small. Only 349 pairs are recorded. However when we set the parameter $k$ from about $0.75m$ to $0.9m$, WPECI is capable of giving the correct answer. If the parameter $k$ can be carefully tuned for each data set, the number of the corrections of the ensemble method can be increased. However it is impossible or may be unfair to do so.

Finally, let us take a look at the unstable example given in the preliminary section.
This data set is recorded as the 83th data set in the above experiment. The setups are the same with that in the Section 2.
%One pair of data is recorded each day. So there are 365 data pairs in this data set.
\begin{figure}[!htb]
	\centering
	\includegraphics[width=305pt, bb=0 20 413 353]{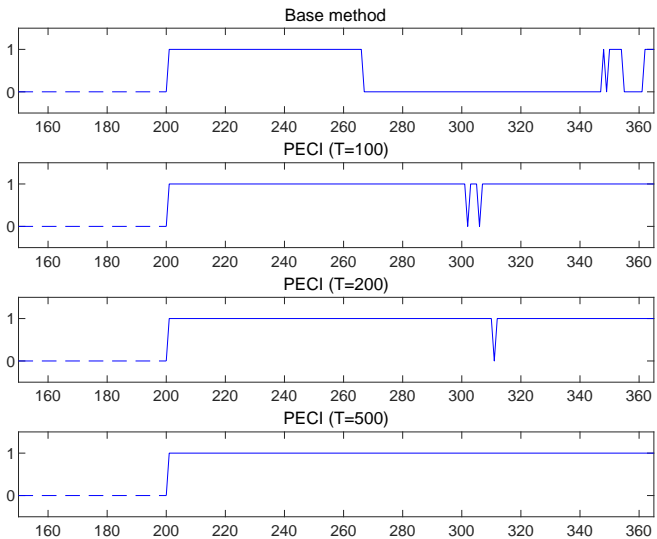}
	\caption{Performance of PECI on an increasing data set with different $T$.}\label{figure_recur}
\end{figure}
%The simulations are done in the way as the data pairs are collected. The first experiment is done with the data pairs collect from the beginning to the 201th day and the second experiment is performed with the data pairs collected to the 202th day etc.
The causal decisions of the base learners are depicted in the top picture of Figure \ref{figure_recur}, which is identical with the Figure \ref{unstable}. It shows that the results of the base method are rather unstable and almost half of the results are incorrect. The second to the fourth pictures show the performances of PECI under $T=100$, $200$, and $500$. The improvements are obvious. Every steps in the simulation $T=500$ gives the same and correct result.
 
%The results of PECI are more stable than the base method. 
\section{Conclusions}
In this paper parallel ensemble based causal direction inference algorithms are proposed. The accuracy rates of the methods are investigated theoretically. Higher correct rates and more stable results are obtained on both the artificial data sets and the real world data sets. In addition, the parallel ensemble framework can be conveniently implemented on parallel computing environments. So the performance of the causal direction inference can be improved with controllable time expenses.
\section*{Acknowledgement}
This work is supported by NSFC-61803337, NSFC-61803338, ZJSTF-LGF18F020011.

\section*{Appendix}

\begin{lemma}\label{lem3}
	(The Hoeffding's inequality)
	Let $r = \sum w_ir_i$, where each of $r_1$, $r_2$,...,$r_N$ is a sum of independent random variables. $r_i$ and $r_j$ need not be mutually independent for $i\ne j$. $w_i$ is positive and $\sum w_i = 1$. If $r_i$ for $i = 1,...,N$ are identically distributed,  the following inequalities holds for any $t>0$.
	\begin{equation}\label{hoeffding1}
	P(r-E(r) \ge t) \le exp(-\frac{2t^2}{\sum_{i=1}^N(max(r_i)-min(r_i))^2}) 
	\end{equation}
	\begin{equation}\label{hoeffding2}
	P(r-E(r) \le -t) \le exp(-\frac{2t^2}{\sum_{i=1}^N(max(r_i)-min(r_i))^2})
	\end{equation}
	%$$P(\bar{X}-E(\bar{X})<t) \le exp(-\frac{2Nt^2}{\sum_{i=1}^N(max(X_i)-min(X_i))^2}) $$
	%$$P(-\bar{X}+E(\bar{X})<t) \le exp(-\frac{2Nt^2}{\sum_{i=1}^N(max(X_i)-min(X_i))^2}) $$ 
	% where $\bar{X}$ is the mean value of $X_i$, and 
	%$$P(|\bar{X}-E(\bar{X})|<t) \le 2exp(-\frac{2Nt^2}{\sum_{i=1}^N(max(X_i)-min(X_i))^2}) $$
	where $E(\cdot)$ denotes the mathematical expectation.
\end{lemma}
The proof of Lemma \ref{lem3} can be found in the section 4 of \cite{hoeffding1994probability}.

Note that the Hoeffding's inequalities with independent assumption are more frequently used in the research field of computer science. However, the inequalities still hold when the random variables are not independent. And this is more in line with the reality in this paper.
The upper bound of the PECI can then be obtained by Lemma 1-3.

%\begin{proof}
\noindent \textbf{Proof of Theorem 1.}
Let $r =\sum_t r_t $, where $r_t$ is defined in the algorithm PECI. $max(r_t) = 1$ and $min(r_t) = -1$.
If the correct causal direction is $x \rightarrow y$, the expectation of $r$ can be obtained by Lemma 2 as
$E(r) = 1\cdot (1-\varepsilon_k)+(-1)\cdot\varepsilon_k = \mathrm{erf}(\sqrt{k-1}\cdot\frac{\mu}{\sqrt{2\sigma^2}}) $,
then the error rate of the ensemble method is 
$$\varepsilon_k^{(T)} =  P(r\ge 0) = P(-r+E(r)\le E(r)).$$
Then the upper bound \eqref{the} can be obtained by using \eqref{hoeffding2}.
Otherwise if the correct causal direction is $y \rightarrow x$, 
$E(r) = -1\cdot (1-\varepsilon_k)+1\cdot\varepsilon_k = -\mathrm{erf}(\sqrt{k-1}\cdot\frac{\mu}{\sqrt{2\sigma^2}}) $
the error rate of the ensemble method is 
$$\varepsilon_k^{(T)} =  P(r\le 0) = P(r) = P(r-E(r)\le -E(r)).$$
Then the upper bound \eqref{the} can be obtained by using \eqref{hoeffding1}.
Note that it is assumed the base learns give results better than random guesses, which means $-E(r)>0$ when the truth is $y \rightarrow x$.
\qed
%\end{proof}

%$$\varepsilon_m = \frac{1}{2}(1-erf(\frac{\sqrt{m}\mu}{\sqrt{2\sigma^2}}))$$
%
%$$erf(t) =\frac{2}{\sqrt{\pi}} \sum_{k=0}^{\infty}\frac{(-1)^k t^{2k+1}}{k!(2k+1)}$$
%
%\begin{align*}
%\varepsilon_{m/N} &= \frac{1}{2}(1-erf(\frac{\sqrt{m}\mu}{\sqrt{N}\sqrt{2\sigma^2}}))\\
%&=\frac{1}{2}(1-\frac{\sqrt{N}}{N-1}erf(\mu/\sigma\sqrt{2}))
%\end{align*}
%$$\varepsilon_{m/N} = \frac{1}{2}(1-erf(\mu/\sqrt{N}\sigma\sqrt{2}))$$
%\end{proof}
%

\bibliographystyle{elsarticle-num} 
\bibliography{mybibfile}

\end{document}